%% file: main.tex
\documentclass{article} 
\usepackage{iclr2019,times}

\usepackage{hyperref}
\usepackage{url}
\usepackage{comment}
\usepackage[flushleft]{threeparttable}
\usepackage{subfig}
\usepackage{algorithm}
\usepackage{enumitem}

\usepackage{amsmath}
\usepackage{graphicx}
\graphicspath{{images/}}
\usepackage{tabularx}
\usepackage{multirow}

\usepackage{comment}
\usepackage{todonotes}
\newcommand{\SideNote}[2]{} 
\renewcommand{\SideNote}[2]{\todo[color=#1,size=\footnotesize]{#2}}

\title{Learning Robust, Transferable Sentence Representations for Text Classification}

\iclrfinalcopy
\author{Wasi Uddin Ahmad$^*$, Xueying Bai$^\mathsection$, Nanyun Peng$^\dagger$, Kai-Wei Chang$^*$   \\
$^*$University of California, Los Angeles \\
$^\mathsection$Stony Brook University, $^\dagger$University of Southern California \\
\texttt{wasiahmad@ucla.edu}, \texttt{xubai@cs.stonybrook.edu} \\ 
\texttt{npeng@isi.edu}, \texttt{kwchang.cs@ucla.edu}
}

\begin{document}

\setlength{\abovedisplayskip}{5pt}
\setlength{\belowdisplayskip}{5pt}
\graphicspath{{images/}}

\maketitle

\begin{abstract}
Despite deep recurrent neural networks (RNNs) demonstrate strong performance in text classification, training RNN models are often expensive and requires an extensive collection of annotated data which may not be available. To overcome the data limitation issue, existing approaches leverage either pre-trained word embedding or sentence representation to lift the burden of training RNNs from scratch. In this paper, we show that jointly learning sentence representations from multiple text classification tasks and combining them with pre-trained word-level and sentence level encoders result in robust sentence representations that are useful for transfer learning. Extensive experiments and analyses using a wide range of transfer and linguistic tasks endorse the effectiveness of our approach.
\end{abstract}

\input{introduction}

\input{relwork}
\input{method}
\input{evaluation}
\input{experiment}

\input{conclusion}


\bibliography{iclr2019}
\bibliographystyle{iclr2019}

\newpage
\appendix
\input{appendix}

\end{document}

%% file: introduction.tex
\section{Introduction}
Recent advances in deep neural networks have demonstrated the capability to build highly accurate models by training on vast amounts of data. 
The efficiency of these techniques comes from their ability to learn an encoder to convert raw inputs into useful continuous feature representations effectively.   
These successes primarily credit to the availability of ample resources, such as an extensive collection of training data.
However, collecting a sufficient amount of manually annotated data is not always feasible, especially for domains requiring expert annotators.


While human annotated data is limited, there are abundant resources that can be used to lift the burden of learning representations from scratch and thus subsidize the requirement of having a large amount of training data.
In the context of modeling natural languages, many successful stories showed that learned representations in both word and sentence levels are transferable to other tasks. 
These pre-trained representations enable us to model many natural language processing (NLP) tasks such as text classification \citep{bailey2018few} and named entity recognition \citep{cherry2015unreasonable} with only a few thousands of examples.

At the word level, pre-trained word embeddings~\citep{mikolov2013distributed,pennington2014glove} encode each word into a continuous vector representation, have been widely used in many applications \citep{seo2016bidirectional,lee2017end,Venugopalan_2017_CVPR,teney2016graph}. 
A few recent methods propose to construct \textit{contextualized word vectors} to address the issue that the meaning of a word should be context-dependent. 
For example, \cite{peters2018elmo} leveraged a large unannotated corpus to train such contextualized word vectors by feeding word sequences into a deep recurrent neural network (RNNs) and generating  representations based on the hidden states of the RNNs correspond to the respective words. 
This results in impressive performances in many NLP applications \citep{lee2017end,peters2017semi,he2017deep}. 

At the sentence level, \cite{conneau2017supervised} showed that an LSTM-based sentence encoder~ \citep{hochreiter1997long} trained on an annotated corpus for natural language inference (NLI) \citep{bowman2015large} can capture useful features that are transferable to a wide range of text classification tasks.
A few follow-up studies \citep{subramanian2018learning,logeswaran2018efficient,cer2018universal} extended the approach by leveraging \emph{large-scale} data and studied how to learn better transferable sentence representations.





However, all the existing approaches considered training word or sentence level representations from scratch. In contrast, we argue that by leveraging pre-trained embeddings/encoders and employing multiple large-scale supervised text classification datasets, we can learn more robust and transferable sentence representations.
The primary research question that we address is how to build robust and transferable representations for sentence classification. 
The key challenges are two folds: 1) how to transfer only salient features by distinguishing generic information from task-specific information when learning an encoder and 2) how to combine representations at word and sentence levels to build strong transferable representations for sentence classification.


To address the first challenge, we propose to leverage multi-task learning (MTL) to jointly train sentence encoders on three large-scale text classification corpora, which cover a variety of domains and two language classification tasks -- textual entailment and question paraphrasing.
We exploit an MTL architecture that learns to separate generic representations from  task-specific representations using adversarial training.
While generic representations capture language-specific information, i.e., language structure, syntax, and semantics that are useful uniformly across a variety of language tasks, the task-specific representations encode domain knowledge that is helpful if the source and transfer tasks are homogeneous.
Our experimental results show that when the shared and task-specific encoders are combined, they become more effective and applicable to a wide range of transfer tasks.

Besides, we combine our MTL-based sentence encoders with another existing sentence encoder \citep{subramanian2018learning} trained with different learning signals, and contextualized word vectors \citep{peters2018elmo}, to build a more robust and transferable sentence encoder.
We evaluate our encoder on 15 transfer \citep{conneau2018senteval} and 10 linguistic probing \citep{conneau2018you} tasks. 
Experimental results demonstrate that our proposed sentence encoder better captures linguistic information and provides a significant improvement over existing transfer learning approaches.

%% file: relwork.tex
\section{Related Work}
Our work is closely related to sentence representation learning, multi-task learning, and transfer learning and we briefly review each of these areas in this section.

\noindent\textbf{$\bullet$ Sentence Representations Learning.}
Training neural networks to form useful sentence representations has become a core component in many machine learning models. 
Learning distributional sentence representations such that they capture the syntactic and semantic regularities has been proposed. These approaches range from models that compose of word embeddings \citep{le2014distributed,arora2016simple,wieting2015towards} to models with complex network architectures \citep{zhao2015self,wang2015learning,liu2016learning,lin2017structured}.
Unsupervised approaches are also proposed in literature by utilizing a large collection of unlabeled text corpora to learn distributional sentence representations.
For example, \cite{kiros2015skip} revised the skip-gram model \citep{mikolov2013distributed} to learn a generic sentence encoder, called SkipThought that is further improved by using layer normalization \citep{ba2016layer}.
Among other closely related works, the technique proposed by \citep{hill2016learning} fell short to SkipThought while \citep{logeswaran2018efficient} showed improvement over skip-thought vectors.

Unlike word embeddings, learning sentence representations in an unsupervised fashion lack the reasoning about semantic relationships between sentences. 
To this end, \cite{conneau2017supervised} proposed to train a universal sentence encoder in the form of a bidirectional LSTM using the {\em supervised} natural language inference data, outperforming unsupervised approaches like SkipThought. 
\cite{subramanian2018learning} propose to build general purpose sentence encoder by learning from a joint objective of classification, machine translation, parse tree generation and unsupervised skip-thought tasks.
Compared to their approach, we propose to utilize multiple text classification datasets by leveraging a multi-task learning approach and combine them with existing contextualized word vectors \citep{mccann2017learned,peters2018elmo} to learn robust and transferable sentence representations.
Recent works \citep{cer2018universal,perone2018evaluation} explored RNN free sentence encoders and evaluated sentence representations learning methods by using a variety of downstream and linguistic tasks.


\noindent\textbf{$\bullet$ Multi-task Learning (MTL).}
Multi-task learning has been successfully used in a wide-range of natural language processing applications, including text classification \citep{liu2017adversarial}, machine translation \citep{luong2015multi}, sequence labeling \citep{rei2017semi}, sequence tagging \citep{peng2017multi}, dependency parsing \citep{peng2017deep} etc.
Recent works \citep{liu2016recurrent,zhang2017generalized,liu2016deep} proposed multi-task learning architectures with different methods of sharing information across the participant tasks.
To facilitate scaling and transferring when a large number of tasks are involved, \cite{zhang2017multi} proposed to embed labels by considering semantic correlations among tasks.
To investigate how much transferable an end-to-end neural network architectures are for NLP applications, \cite{mou2016transferable} propose to use multi-task learning on sentence classification tasks. 
In contrast to these prior work, we aim to learn a universal sentence encoders via multi-task learning that are transferable to a wide range of heterogeneous tasks. 


\noindent\textbf{$\bullet$ Transfer Learning.}
Transfer learning stores the knowledge gained from solving source tasks (usually with abundant annotated data), and apply it to other tasks (usually suffer from insufficient annotated data to train complex models) to combat the inadequate supervision problem. 
It has become prevalent in many computer vision applications \citep{sharif2014cnn,antol2015vqa} where image features were trained on ImageNet \citep{deng2009imagenet}, and applications where word vectors \citep{pennington2014glove,mikolov2013distributed} were trained on large unlabeled corpora.
Despite the benefits of using pre-trained word embeddings, many NLP applications still suffer from lacking high quality generic sentence representations that can help unseen tasks.
In this work, we combine sentence representations learned using MTL and contextualized word vectors to obtain more robust sentence representations that transfer better. 

%% file: method.tex
\section{Sentence Representations Learning}


Our goal is to leverage available text copora and existing sentence and word encoders to build a universal sentence encoder. 
In the following, we first define the sentence encoder and then describe a multi-task learning approach that learns sentence representations jointly on multiple text classification tasks.
Then we discuss how to combine the learned sentence representations with the existing sentence and contextualized word vectors.


\subsection{Sentence Encoder}
A typical text classification model consists of two parts: a representation learning component, also known as encoders that convert input text sequences into fixed-size vectors, and a classifier component that takes the vector representations and predicts the final class labels.
The encoder is usually realized by a high complexity neural network architecture and requires a large amount of data to train, as opposed to the classifier which is generally simple (e.g., a linear model).
When enough training examples are provided, the encoder and the classifier can be trained jointly from scratch in an end-to-end fashion.\footnote{In this case, the classifier is the last layer in the network architecture.} However, when data is insufficient, this approach is unfeasible. Instead, we can pre-train the encoder on other tasks (a.k.a source tasks) and transfer the learned encoder to the target task.
In this case, we only require a few labeled examples to train the low-complexity classifier on top of the pre-trained encoder. We discuss how to build the pre-train encoder in below.

We follow  \citep{conneau2017supervised} to build a transferable encoder based on an one layer bidirectional LSTM with max pooling (BiLSTM-max).
Formally, given a sentence with $T$ words, $[w_{1}, w_{2},..., w_{T}]$, the encoder first runs two LSTM models on input text from both directions. 
\begin{equation}
\label{eqn:bilstm}
\begin{aligned}
    \overrightarrow{h}_t = LSTM(\overrightarrow{h}_{t-1}, w_{t}), \ 
    \overleftarrow{h}_t  = LSTM(\overleftarrow{h}_{t+1}, w_{t})
\end{aligned}
\end{equation}
and $h_t = [\overrightarrow{h}_t, \overleftarrow{h}_t] \in R^{2d}$ is the $t$-th hidden vectors in BiLSTM, $d$ is the dimensionality of the LSTM hidden units.
To form a fixed-size vector representation of variable length sentences, the maximum value is selected over each dimension of the hidden units:
\begin{equation}
\label{eqn:maxpooling}
    s_{j} = \max\nolimits_{t \in [1,\ldots,T]} \ h_{j, t},\ j = 1, ..., d,
\end{equation}
where $s_{j}$ is the $j$-th element of the sentence embedding $s$.

For some transfer tasks (e.g., textural entailment and similarity measuring), the goal is to predict the relationship between two sentences. Therefore, the input involves two sentences  $(s_{1}, s_{2})$. We generate the representation of input instances by $[s_{1}, s_{2}, s_{1} - s_{2}, s_{1} \odot s_{2}]$
where $\odot$ denotes the element-wise multiplication, and $[\cdot, \cdot]$ denotes vector concatenation. 


\noindent\textbf{Multi-task learning.} Multi-task learning was shown efficient in many text classification tasks.
However, its effectiveness in learning transferable sentence representations is comparably less studied.
In this paper, we investigate the utility offered by several large-scale text classification tasks.
We show that learning signals from various text classification tasks results in  robust and transferable sentence representations.
Inspired by \citep{liu2017adversarial},
we study two variants of shared-private (SP) multi-task learning (MTL) frameworks.
The shared-private MTL framework maintains private and shared encoders with their own task-specific layers to encourage task-specific and generic features being learned by the private and shared encoders, respectively. 
In this study, we design one private BiLSTM-max sentence encoder for each task, and one shared BiLSTM-max encoder for all the tasks to capture task-dependent and generic features, respectively. Sentence embeddings produced by private and shared encoders are concatenated to form the final sentence representations. In this way, the shared encoder provides task-independent information, while the private encoders are helpful when the target task is proximity to some source tasks.
Formally, for any sentence in a given task $k$, its shared 
representation $s^{k}_{s}$ and private representation $s^{k}_{p}$ can be computed using Eq. \eqref{eqn:bilstm} -- \eqref{eqn:maxpooling},
and they are concatenated to construct the sentence embedding: $s^{k}=[s^{k}_{s}, s^{k}_{p}]$.

\noindent\textbf{Adversarial training.}
Ideally, we want the private encoders to learn only task-specific features, and the shared encoder to learn generic features.
To achieve this goal, we adopt the adversarial training strategy proposed by \cite{liu2017adversarial} to introduce a discriminator on top of the shared BiLSTM-max sentence encoder.
The goal of the discriminator, $D$ is to identify which task an encoded sentence $s^k$ comes from, and the adversarial training requires the shared sentence encoder to generate representations that can ``fool'' the discriminator. In this way, the shared encoder is forced not to carry task-related information.
The discriminator is defined as,
\begin{align*}
    D(s^k) = \mbox{\it softmax}(W s^k + b),
\end{align*}
where $W \in R^{d \times d}$ and $b \in R^d$ are model parameters.
Optimizing the adversarial loss,
\begin{equation*}
    L_{adv} = \min_{\theta_{E}} \Bigg(\max_{\theta_{D}}\big(\sum_{k=1}^{K}\sum_{i=1}^{N_k} d_{i}^{k}\log[D(E(s))]\big)\Bigg)
\end{equation*}
has two competing goals: the discriminator tries to maximize the classification accuracy (inside the parentheses), and the sentence encoder tries to confuse it (and thus minimize the classification accuracy).
$E$ and $D$ represents the shared sentence encoder and the discriminator respectively and $\theta_E$ and $\theta_D$ are the model parameters of $E$ and $D$. $d_i^{k}$ denotes the ground-truth label indicating the type of the current task.
To encourage the shared and private encoders to capture different aspects of the sentences,
the following term is added.
\begin{align*}
    L_{diff} = \sum\nolimits_{k=1}^{K} \left \|  H_{s}^{k^\top} H_{p}^{k} \right \|_{F}^{2}
\end{align*}
where $\left\|\cdot\right\|_{F}^{2}$ is the squared Frobenius norm. 
Here, $H_{s}^{k}$ and $H_{p}^{k}$ are matrices where rows are the hidden vectors (see Eq. (1)) generated by the shared and private encoders given an input sentence of task $k$.
The final loss function is a weighted combination of three parts:
\begin{align*}
    L & = L_{multi-task}+\beta L_{adv} + \gamma L_{diff}
\end{align*}
where $\beta$ and $\gamma$ are hyper-parameters, $L_{multi-task}$ refers to a simple summation over the cross entropy loss for each task.
We tune $\beta$ and $\gamma$ in the range $[0.001, 0.005, 0.01, 0.05, 0.1, 0.5]$ 
and present the best $\beta$ and $\gamma$ values in table \ref{table:param_tuning} (provided in the appendix) for different multi-task learning settings.

\subsection{Unifying Sentence Embeddings and Contextualized Vectors}
Existing studies \citep{subramanian2018learning,peters2018elmo} leverage large amount of data to train sentence or word representations. 
However, sometimes it is impractical to assume there is an access to these large-scale data or computation resources. In these circumstances, we can combine existing encoders in a post-processing step to leverage the humongous data sources. 
In this work, we show that by combining our MTL based sentence encoder with an existing sentence encoder \citep{subramanian2018learning} and a contextualized word representation \citep{peters2018elmo} encoder, we achieve state-of-the-art transfer performance on a wide variety of text classification tasks.

The contextualized word vectors refer to the hidden states generated by a BiLSTM (as in Eq. \eqref{eqn:bilstm}) given a sequence of words (sentences) as inputs.
To form a fixed size sentence representation from the contextual word vectors, we apply average pooling\footnote{We tried max pooling, but it consistently performed more inferior compared to average pooling.}.
Although \cite{peters2018elmo} suggested learning the weights of the contextual word vectors, we do not learn any additional weights because we consider scenarios when there is no training example available to learn such weights.
To get a universal sentence representation, we concatenate the sentence embeddings provided by our MTL based encoders, an existing sentence encoder \citep{subramanian2018learning} trained with another set of tasks, and the fixed size vector constructed from contextual word vectors \citep{peters2018elmo}.
We will investigate more effective ways to combine multiple representations in our future work.

%% file: evaluation.tex
\section{Experiments}
In this section, we first show that our proposed sentence encoder can achieve state-of-the-art transfer performances. 
Then we demonstrate that combining the multi-task trained sentence representations with other sentence and word vectors yield better universal sentence representations.
To better understand our results, we provide a thorough ablation study and confirm our combined encoder can learn robust and transferable sentence representations.

\subsection{Experimental Setup}
 We use three large-scale textual entailment and paraphrasing tasks to train sentence encoders with multi-task learning, and combine these with an existing sentence encoder and contextualized word embeddings to compose the final sentence representations. We test the generalizability of the sentence embeddings on fifteen transfer tasks.
A detailed description of the source and transfer tasks are presented in table \ref{table:data_stat} in the appendix.
In addition, we perform a quantitative analysis using ten probing tasks to show what linguistic information is captured by our proposed sentence encoders.

\noindent\textbf{Source tasks.} The first two source tasks are natural language inference (NLI) which determines whether a natural language hypothesis can be inferred from a natural language premise. 
We consider the SNLI \citep{bowman2015large} and the Multi-Genre NLI (MNLI) \citep{williams2017broad} which consist of sentence pairs, manually labeled with one of the three categories: entailment, contradiction and neutral.
Following \cite{conneau2017supervised}, we also conduct experiments that combine SNLI and MNLI datasets, which is denoted as AllNLI.
The second task is the Quora question paraphrase (QQP)\footnote{https://www.kaggle.com/quora/question-pairs-dataset} detection based on a dataset of 404k question pairs. We use the Quora dataset split as that in \cite{wang2017bilateral}. 
We present and discuss the source task performances in appendix \ref{sec:eval_src_tasks}.


\noindent\textbf{Transfer and probing tasks.}
We evaluate the sentence encoders on fifteen transfer and ten probing tasks using the SentEval toolkit \citep{conneau2018senteval}.
Among the transfer tasks, six are text classification tasks for sentiment analysis (MR, SST), question-type (TREC), product reviews (CR), subjectivity/objectivity (SUBJ) and opinion polarity (MPQA). 
Rest of the transfer tasks (SICK-E, SICK-R, MRPC, STSB, and STS12--16) are semantic relatedness and textual similary tasks.
We test our sentence encoder on capturing linguistic (surface, syntactic, and semantic) information using the ten probing tasks suggested in \citep{conneau2018you}.


\noindent\textbf{Hyper-parameter tuning.} We carefully tune the parameters and report the testing performance with best parameters. 
We use SGD with an initial learning rate of $0.1$ and a weight decay of $0.99$. 
At each epoch, we divide the learning rate by $5$ if the development accuracy decreases. 
We use mini-batches of size $128$ and training is stopped when the learning rate goes below the threshold of $10^{-5}$. 
For the task-specific classifier, we use a multi-layer perceptron with $1$ hidden-layer of $512$ hidden units. 
We consider the range $[256, 512, 1024, 2048]$ for the number of hidden units in BiLSTM and found $2048$ results in best performance. 
We use $300$ dimensional GloVe word vectors \citep{pennington2014glove} trained on $840$ billions of tokens as fixed word embeddings.


\input{table/table4.tex}

\input{table/table5.tex}


%% file: table/table4.tex
\begin{table}[t]
\centering
\resizebox{\linewidth}{!}{%
\begin{tabular}{@{}l@{\hskip 0.2in} c@{\hskip 0.15in} c@{\hskip 0.15in} c@{\hskip 0.08in} c@{\hskip 0.1in} c@{\hskip 0.1in} c@{\hskip 0.08in} c@{\hskip 0.05in} c@{}}
\hline
Model Type & MR   & CR   & SUBJ & MPQA & SST  & TREC & SICK-E & MRPC  \\ 
\hline 
\multicolumn{9}{l}{1.Unsupervised sentence representations learning} \\ 
1.1. FastSent \citep{hill2016learning} & 70.8 & 78.4 & 88.7 & 80.6 & - & 76.8 & - & 72.2/80.3 \\
1.2. SkipThought \citep{kiros2015skip} & 76.5 & 80.1 & 93.6 & 87.1 & 82.0 & 92.2 & 82.3 & 73.0/82.0 \\
1.3. USE (Transformer) \citep{cer2018universal} & 81.4 & 87.4 & 93.9 & 87.0 & 85.4 & 92.5 & - & - \\
1.4. Byte mLSTM \citep{radford2017learning} & \textbf{86.9} & \textbf{91.4} & 94.6 & 88.5 & - & - & - & 75.0/82.8 \\
\hline
\multicolumn{9}{l}{2. Supervised sentence representations learning} \\ 
2.1. InferSent \citep{conneau2017supervised}   & 81.6 & 85.9 & 92.4	& 90.4 & 85.3 & 87.0 & 85.6 & 75.5/82.2 \\
2.2. GenSen \citep{subramanian2018learning}   & 82.7 & 87.6 & 94.1 & 91.1 & 83.5 & 93.0 & 87.8  & \textbf{78.0/84.2} \\
2.3. Sent2vec (SP) \textbf{(this paper)} & 81.7	& 86.3	& 93.7	& 90.8	& 86.2	& 89.4	& 86.0 &  75.2/82.2 \\
2.4. Sent2vec (ASP) \textbf{(this paper)} & 82.3 & 86.3 & 93.5 & 90.8 & 84.2 & 89.4 & 87.1 & 76.3/83.2 \\
\hline 
\multicolumn{9}{l}{3. Contextualized word vectors} \\ 
3.1. CoVe \citep{mccann2017learned} & 75.3 & 77.2 & 89.1 & 88.3 & 80.3 & 85.6 & 78.7 & 70.3/81.6 \\
3.2. ELMo \citep{peters2018elmo} & 81.2 & 84.2 & 95.0 & 90.2 & 87.2 & 92.8 & 81.4 & 76.0/82.3 \\
\hline 
\multicolumn{9}{l}{4. Sentence representations combined with contextualized word vectors \textbf{(this paper)}} \\ 
4.1.  Sent2Vec (ASP) + ELMo &  84.5 & 87.7 & 95.6 & \textbf{91.7} & \textbf{88.7} & 93.8 & 86.3 & 77.0/84.2 \\
4.2.  GenSen + ELMo & 85.0 & 88.0 & 95.8 & 91.5 & 85.9 & \textbf{94.8} & 87.0 & 77.4/83.5 \\
4.3. Sent2vec (ASP) + GenSen + ELMo & 85.9 & 88.3 & \textbf{96.2} & 91.6 & 87.7 & 94.6 & \textbf{88.4} & 77.2/84.4 \\
\hline 
\multicolumn{9}{l}{5. Approaches trained from scratch on the tasks} \\
5.1. AdaSent \citep{zhao2015self} & 83.1 & 86.3 & 95.5 & 93.3 & - & 92.4 & - & -  \\
5.2. TF-KLD \citep{ji2013discriminative} & - & - & - & - & - & - & - & 80.4/85.9  \\
5.3. Illinois LH \citep{lai2014illinois} & - & - & - & - & - & - & 84.5 & - \\
5.4. BLSTM-2DCNN \citep{zhou2016text} & 82.3 & - & 94.0 & - & 89.5 & 96.1 & - & - \\ 
\hline
\end{tabular}
}
\caption{Evaluation of sentence representations on a set of 8 tasks using a logistic regression classifier. 
``SP'' and ``ASP'' in row 2.3 and 2.4 refers to the shared-private and adversarial shared private multi-task learning models.
Values indicate the accuracy (accuracy/F1 for MRPC) for the test sets and bold-faced values denote the best \emph{transfer} performances. 
We employ an averaging bag-of-words technique to form sentence embeddings, using features from all three layers of ELMo.
}
\label{table:transfer_task_results}
\end{table}

%% file: table/table5.tex
\begin{table}[ht]
\centering
\small
\begin{tabular}{@{}l c c c c c c c@{}}
\hline
\multirow{2}{*}{Model Type} & \multirow{2}{*}{SICK-R} & \multirow{2}{*}{STSB} & \multicolumn{5}{c}{Semantic Textual Similarity (STS)} \\
 & & & 2012 & 2013 & 2014 & 2015 & 2016 \\
\hline 
InferSent \citep{conneau2017supervised}  & 0.884 & 0.756 & \textbf{0.61} & 0.56 & 0.68 & 0.71 & 0.71  \\
GenSen \citep{subramanian2018learning} & 0.888 & 0.786 & \textbf{0.61} & 0.54 & 0.65 & 0.74 & 0.67 \\
USE (Transformer) \citep{cer2018universal} & 0.860 & \textbf{0.814} & \textbf{0.61} & \textbf{0.64} & \textbf{0.71} & \textbf{0.74} & 0.74 \\
\hline 
CoVe \citep{mccann2017learned} & 0.808 & 0.745 & 0.50 & 0.40 & 0.63 & 0.54 & 0.61 \\
ELMo \citep{peters2018elmo} & 0.813 & 0.661 & 0.55 & 0.49 & 0.62 & 0.67 & 0.62 \\
\hline 
Sent2vec (SP) \textbf{(this paper)} & 0.887 & 0.752 & 0.55 & 0.56 & 0.64 & 0.67 & 0.72 \\ 
Sent2vec (ASP) \textbf{(this paper)} & 0.888 & 0.738 & 0.57 & 0.59 & 0.67 & 0.71 & 0.74 \\
\hline 
Sent2vec + ELMo \textbf{(this paper)}  & 0.888 & 0.720 & 0.60 & 0.55 & 0.66 & 0.70 & 0.70 \\
Sent2vec + GenSen + ELMo \textbf{(this paper)}  & \textbf{0.895} & 0.753 & \textbf{0.61} & 0.55 & 0.67 & 0.71 & \textbf{0.75} \\
\hline
\end{tabular}
\caption{Transfer evaluation of the semantic relatedness and textual similarity tasks.
``SP'' and ``ASP'' in block 3 refers to the shared-private and adversarial shared private multi-task learning models.
In block 4, we use ASP setting for Sent2vec.
We use features from the top layer of the ELMo to produce sentence embeddings. 
Values indicate the Pearson correlation coefficient for the test sets and bold-faced values denote the best performance across all the models.
}
\label{table:unsup_task_comparison}
\vspace{-2mm}
\end{table}

%% file: experiment.tex


\subsection{Evaluation on Transfer Tasks}
\label{subsec:transfer_eval}
We benchmark the performances of our proposed encoders on fifteen transfer tasks in comparison with several baselines including unsupervised models, sentence encoders, contextualized word encoders and supervised models. 
Table \ref{table:transfer_task_results} and \ref{table:unsup_task_comparison} summarize the results. 
From the block 4 of table \ref{table:transfer_task_results}, we see our combined encoders achieve best performance on 5/8 transfer tasks, which demonstrates the efficacy of our proposed unified sentence encoder. 

We further analyze the performance of our proposed combined encoders from two aspects: the improvement achieved by multi-task learned sentence encoders and the efficiency of the combination.
From block 2 of table \ref{table:transfer_task_results}, we see the MTL based sentence encoders, Sent2vec outperforms the single task based sentence encoder InferSent on 7 out of 8 transfer tasks (comparing row 2.1 with 2.3--2.4). Sent2Vec also provides  competitive performance on the other 7 tasks as shown in table \ref{table:unsup_task_comparison}.
The results demonstrate that learning from multiple tasks helps to capture more generalizable features that are suitable for transfer learning.
In addition, when using adversarial training, we observe improvements in 9 out of 15 transfer tasks comparing to the non-adversarial setting (see table \ref{table:transfer_task_results} and \ref{table:unsup_task_comparison}). 
To investigate the advantages of adversarial training, we provide a detailed comparison between the shared and private encoders with and without adversarial training in appendix \ref{sec:sp_encoder_comp}.

Combining sentence encoders and contextualized word vectors improves the transfer learning performance significantly (comparing row 4.3 to 2.2, 2.4, and 3.2 in table \ref{table:transfer_task_results}). 
First, we see from table \ref{table:unsup_task_comparison}, when there are no training examples available for tasks (STS12--16 and STSB tasks), sentence embeddings (blocks 1 and 3) perform better than contextualized vectors (block 2).
The result demonstrates the necessity of learning generic sentence representations so that they can be directly used (without training) in transfer tasks.
Although Sent2vec outperforms ELMo on 12/15 tasks (comparing row 2.4 to 3.2 in table \ref{table:transfer_task_results} and row ELMo to Sent2vec (ASP) in table \ref{table:unsup_task_comparison}), it fell short to GenSen on most of the transfer tasks since GenSen is trained using 124M sentence pairs while Sent2vec is trained on 1.4M pairs.
Because training an encoder on massive datasets require large computational resources, the direct utilization of GenSen through a post-processing step can bring benefits from large resources in a computational efficient way.
Second, contextualized vectors (ELMo) perform better in specific tasks like SST. Also contextualized vectors are shown capable to capture specific linguistic properties (more details in the analysis part). These indicate that utilizing contextualized vectors may help learn better sentence representations. 
As a result, in the block 4 of table \ref{table:transfer_task_results}, when Sent2vec, GenSen, and ELMo are combined, we observe a significant improvement on 4 out of 8 tasks (MR, CR, SST, and SICK-E) over its individual components and competitive performance on the other tasks, which confirms the efficiency of the combination.

\begin{figure}
\centering
\subfloat[\label{fig:rsub1}]
{
\includegraphics[width=.45\linewidth]{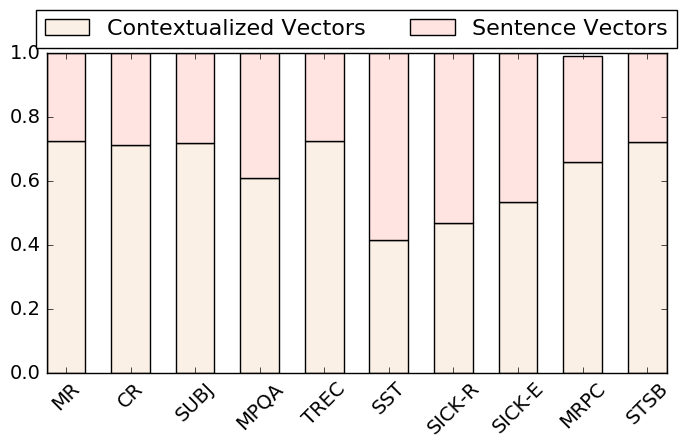}
}
\subfloat[\label{fig:rsub2}]
{
\includegraphics[width=.45\linewidth]{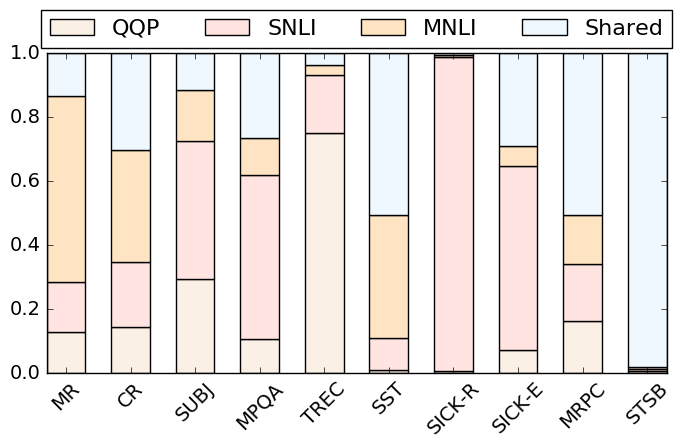}
}
\vspace{-2mm}
\caption{
(a) Weights learned by transfer tasks for contextualized vectors (ELMo, CoVe), and sentence vectors which refer to a concatenation of all the private and shared encoders of the adversarial shared-private multi-task model.
(b) Weights learned by the transfer tasks for private (task-specific) and shared encoders of adversarial shared private multi-task model. }
\vspace{-2mm}
\label{taskvsencoder}
\end{figure}

\subsection{Analysis}
\noindent\textbf{Impact of sentence embeddings and contextualized word vectors.}
To analyze the contributions of our proposed sentence encoder and the sentence representations learned from contextualized word vectors (CoVe, ELMo) in a combined encoder during transfer learning, we design a classifier with a different network architecture. The classifier first generates predicted class probabilities based on a softmax layer using each sentence representations as input. Then the predictions are combined by a pooling layer with a weight parameter for each encoder. By investigating the learned weights in the pooling layer, we can understand which encoder contributes the most.
The learned weights are shown in Figure \ref{taskvsencoder}(a). 
Although contextualized vectors have higher weights in 7/10 tasks, sentence vectors have $>20\%$ contributions for each task and play dominant roles in tasks like SST and SICK-R. 
As we have shown in table \ref{table:transfer_task_results}, the combined encoder performs better than individual encoders (comparing row 4.1 to 2.4, 3.1, 3.2), indicating the contribution of the sentence and contextual word encoders are quite complementary.

\noindent\textbf{Impact of source tasks on transfer tasks.}
To understand the influences of the source tasks on the transfer tasks, 
We conduct a similar analysis as in Figure \ref{taskvsencoder}(a) and show the learned weights assigned for the private (task-specific) and shared (generic) encoders in the ASP model in Figure \ref{taskvsencoder}(b). 
In general, for target tasks that are similar to the source tasks, the private  encoders get higher weights,  otherwise, the shared encoder is better. The combination of the shared and private  encoders enables the transfer task to choose the best combination, thus achieved the best results. 
Most of the transfer tasks assign large weights on the SNLI task-specific and low weights on the QQP task-specific encoder, which explains why representations learned on SNLI are especially efficient for transfer tasks as noted in \citep{conneau2017supervised}. 
Besides, with adversarial training enforced, the shared encoder gets a lower weight from most of the transfer tasks than non-adversarial training (see figure \ref{shared_private_encoder_wgts} in the appendix) demonstrating the efficiency of  adversarial training to separate generic and task-specific  representations. 


\input{table/table8.tex}


\noindent\textbf{Probing for linguistic properties.}
To understand what linguistic properties are captured by our proposed sentence encoders and the unified encoders, we conduct experiments on probing tasks proposed in \cite{conneau2018you}. 
Results are provided in table \ref{table:probing_tasks}. 
Sentence representations show superiority in hard semantic tasks like SOMO and CoordInv, while contextualized word vectors perform better on embedding surface and syntactic properties.
Moreover, MTL encoders (Sent2Vec) outperform both contextualized word vectors (ELMo) and the sentence encoder trained on single task (Infersent) on hard semantic tasks. 
The unified sentence encoder that combines both the sentence and contextual word representations  captures most of the linguistic properties.

\begin{figure}
\centering
\subfloat
{
\includegraphics[width=.42\linewidth]{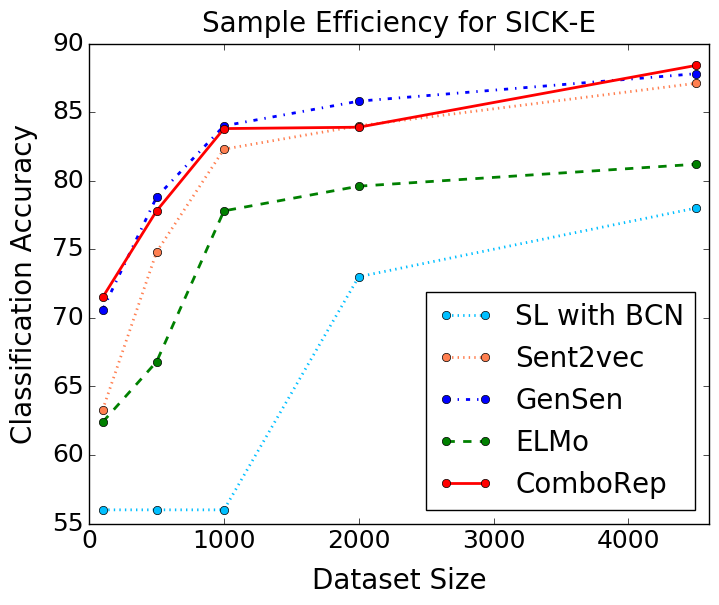}
}
\subfloat
{
\includegraphics[width=.448\linewidth]{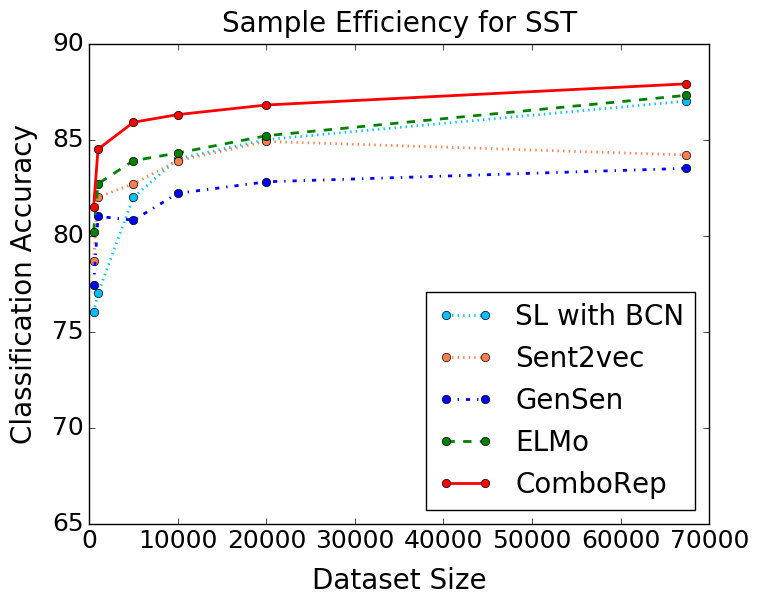}
}
\caption{Comparing test performances of supervised learning (using BCN), word (ELMo) and sentence level representations (Sent2vec, GenSen) and their combination (ComboRep refers to Sent2vec + GenSen + ELMo) on SST and SICK-E tasks as the training dataset size is varied. 
}
\label{sample_sicke}
\vspace{-2mm}
\end{figure}

\noindent\textbf{Impact of training data.}
Finally, we study the sample efficiency of the sentence and contextualized word encoders, as well as a strong supervised learning baseline, BCN \citep{mccann2017learned}, training from scratch on SST and SICK-E tasks.
The results are shown in Figure \ref{sample_sicke}.
We see that the transfer setting have better sample efficiency especially when the training data is limited ($< 5000$ samples). 
Besides, our proposed sentence encoder Sent2vec outperforms the GenSen encoder on the SST tass but it fall short on the SICK-E task.
We show that the combined sentence encoder has higher sample efficiency (can be trained with less labeled examples) than individual ones.
We compare single and multi-task sentence encoders by varying the dataset size and present the results in the appendix \ref{sec:stl_vs_mtl}.

%% file: table/table8.tex
\begin{table*}[t]
\centering
\resizebox{\linewidth}{!}{%
\begin{tabular}{@{}l@{\hskip 0.07in} c@{\hskip 0.07in} c@{\hskip 0.07in} c@{\hskip 0.07in} c@{\hskip 0.07in} c@{\hskip 0.07in} c@{\hskip 0.07in} c@{\hskip 0.07in} c@{\hskip 0.07in} c@{\hskip 0.07in} c@{}}
\hline
Model Type & SentLen  & WC  & TreeDepth & TopConst & BShift & Tense & SubjNum & ObjNum & SOMO & CoordInv  \\ 
\hline 
InferSent & 84.0 & 90.5 & 38.6 & 47.3 & 62.3 & 87.1 & 85.9 & 81.5 & 59.8 & 68.5 \\
GenSen & 93.9 & \textbf{96.7} & 44.0 & 64.1 & 75.5 & 90.0 & 90.1 & 90.3 & 53.6 & 69.0 \\
USE (Transformer) & 79.8 & 54.2 & 30.5 & \textbf{68.7} & 60.5 & 86.2 & 77.8 & 74.6 & 58.5 & 58.2 \\
CoVe & 88.4 & 19.0 & 41.2 & 39.1 & 68.8 & 86.2 & 84.6 & 85.4 & 50.8 & 61.3 \\
ELMo & \textbf{95.3} & 76.0 & 42.6 & 50.4 & 85.1 & 89.6 & 91.4 & 89.1 & 59.6 & 67.8  \\
\hline
Sent2Vec (SP) & 87.3 & 88.1 & 41.9 & 51.9 & 62.8 & 88.1 & 87.7 & 83.9 & \textbf{60.6} & 71.1 \\
Sent2Vec (ASP) & 88.0 & 85.5 & 41.0 & 53.2 & 53.1 & 88.3 & 87.3 & 83.3 & 49.9 & 70.8 \\
\hline
\begin{tabular}{@{}p{2.6cm}}
    Sent2Vec (ASP) + \\ ELMo + GenSen
\end{tabular}  & 91.0 & 86.2 & \textbf{45.2} & 59.2 & \textbf{85.4} & \textbf{91.0} & \textbf{92.4} & \textbf{91.7} & 49.9 & \textbf{73.3} \\
\hline
\end{tabular}
}
\caption{Probing task accuracies with MLP as the classifier. 
For ELMo, the same bag-of-words averaging technique is employed as used for the downstream transfer tasks.
When ELMo is combined with Sent2Vec and GenSen, features only from the top layer are used to fit in single GPU (Titan X).
Bold-faced values denote the best results across the board. 
}
\label{table:probing_tasks}
\vspace{-2mm}
\end{table*}

%% file: conclusion.tex
\section{Conclusion}
In this paper, we propose to leverage available large-scale text classification datasets and existing word and sentence encoding models to learn a universal sentence encoder. 
We utilize multi-task learning (MTL) to train sentence encoders that learn both generic and task-specific sentence representations from three heterogeneous text classification corpora. Experiments show that the MTL trained representations outperforms sentence encoders trained on singe task on a variety of transfer sentence classification tasks. 
We then further combine these sentence encoders with an existing multi-task pre-trained sentence encoder (with a different set of tasks) and a contextualized word representation learner. Our proposed unified sentence encoder yields significant improvements over the state-of-the-art sentence representations on transfer learning tasks.
Extensive comparisons and thorough analysis using 15 transfer datasets and 10 linguistic probing tasks endorse the robustness of our proposed universal sentence encoder.


%% file: appendix.tex
\input{table/table1.tex}
\input{table/table9.tex}

\begin{figure}
\centering
\subfloat[\label{fig:rsub1}]
{
\includegraphics[width=.45\linewidth]{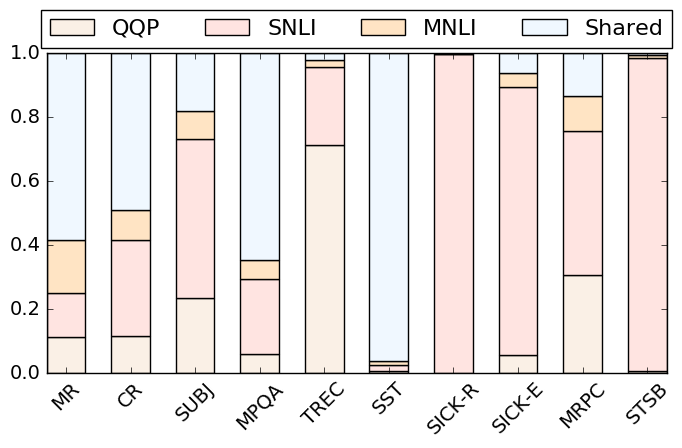}
}
\subfloat[\label{fig:rsub2}]
{
\includegraphics[width=.45\linewidth]{images/mtl_encoders.png}
}
\vspace{-2mm}
\caption{
Weights learned by the transfer tasks for private (task-specific) and shared encoders of shared private multi-task model (a) with and (b) without adversarial training.}
\label{shared_private_encoder_wgts}
\end{figure}

\section{Evaluation on Source Tasks}
\label{sec:eval_src_tasks}
In this section, we discuss the performance of the shared-private multi-task learning (MTL) frameworks on different combinations of QQP, SNLI and MNLI datasets as the {\em source tasks}.
We concatenate the representations generated by the shared and private encoders to form sentence embeddings.
The results are presented in table \ref{table:source_task_comparison}.
We compare the performance of MTL with the models trained on single tasks as in \citep{conneau2017supervised}.
Table~\ref{table:source_task_comparison} shows that learning from multiple tasks performs better than learning from a single task.
However, to our surprise, the adversarial training does not {\em always} excel on source tasks but we show in the transfer evaluation that adversarial training boosts the transfer learning performance.


\input{table/table11.tex}

\input{table/table12.tex}

\input{table/table7.tex}

\section{Singel-task vs. multi-task learning with varying training data}
\label{sec:stl_vs_mtl}
When we compare MTL to STL for transfer
learning, one fundamental question that arises is, does improvement in transfer learning via MTL only come because of having more annotated data? 
Comparing the performance of AllNLI in a single task setting and \{SNLI, MNLI\} in the multi-task settings in table \ref{table:appendix_comparison}, we observe significant improvement in 7/10 tasks. 
In both settings, the amount of training data is the same.
To verify the hypothesis that the improvements in transfer learning do not solely come from having more annotated data, we design an experiment that samples equal amount of data (225k training examples) from SNLI and QQP to match the size of full SNLI dataset.
We found 0.26\% average improvement in transfer tasks compared to single task learning (STL) on the SNLI dataset. 
With full SNLI and QQP dataset, we observe a larger (0.69\% on average) improvement in transfer tasks compared to STL on SNLI dataset.
The first row of table~\ref{table:wpb} shows that MTL is beneficial in this setting and the second row demonstrates that with additional data, MTL achieves larger improvements.

\section{Private Encoders vs. Shared Encoder}
\label{sec:sp_encoder_comp}
To verify our hypothesis that shared encoder learns generic features that are more suitable for transfer learning and with adversarial training enforced, the shared encoder becomes more effective; we provide a detailed comparison of the private and shared encoders in table \ref{table:sp_comparison}.
However, by concatenating the shared and task-specific representations, we can achieve better transfer performance which indicates that transfer tasks also get benefited from task-specific features, specially when the source and transfer tasks are homogenous (more details are provided in the ablation analysis).

\input{table/table10.tex}


%% file: table/table1.tex
\begin{table}[ht]
\centering
\small
\begin{tabular}{l|l|l}
\hline
Tasks & $\beta$ & $\gamma$ \\ 
\hline
QQP and SNLI & 0.01 & 0.05 \\
SNLI and MNLI & 0.005 & 0.001 \\
QQP and AllNLI & 0.01 & 0.05 \\
QQP, SNLI and MNLI & 0.005 & 0.001 \\
\hline
\end{tabular}
\caption{Best $\beta$ and $\gamma$ values for adversarial shared private model on different set of tasks.}
\label{table:param_tuning}
\end{table}

%% file: table/table9.tex
\begin{table}[ht]
\centering
\small
\begin{tabular}{l|r|r|p{2.3cm}|c} 
\hline
Name & \multicolumn{1}{c|}{N} & \multicolumn{1}{c|}{V} & Task & C \\
\hline 
\hline
\multicolumn{4}{l}{Binary and multi-class classification tasks} \\ \hline
MR	    & 11k & 20.3k & sentiment & 2 \\ 
CR	    & 4k & 5.7k & product review & 2 \\ 
SUBJ	& 10k & 22.6k & subj/obj & 2 \\ 
MPQA	& 11k & 6.2k & opinion  & 2 \\ 
SST 	& 70k & 17.5k & sentiment  & 2 \\ 
TREC	& 6k & 9.7k & question-type & 6 \\ 
\hline 
\hline
\multicolumn{4}{l}{Recognizing textual entailment tasks} \\ \hline
SNLI$^{\dag}$  & 560k & 42.7k &  entailment & 3 \\ 
Multi-NLI$^{\dag}$ & 433k & 102.7k &  entailment & 3 \\ 
SICK-E 	& 10k & 2.4k &  entailment & 3 \\ 
\hline 
\hline
\multicolumn{4}{l}{Paraphrase identification tasks} \\ \hline
QQP$^{\dag}$ & 404k & 127.5k &  paraphrasing & 2 \\
MRPC 	& 5.8k & 19.5k &  paraphrasing & 2 \\
\hline 
\hline
\multicolumn{4}{l}{Semantic textual similarity tasks} \\ \hline
SICK-R 	& 10k & 2.4k &  similarity & 0 -- 5 \\
STSB  & 8.6k & 15.9k & similarity & 5 \\
STS-12 	& 399 & 735 &  similarity & 0 -- 5\\
STS-13 	& 561 & 1.6k &  similarity & 0 -- 5\\
STS-14 	& 750 & 3.8k &  similarity & 0 -- 5\\
STS-15 	& 750 & 1.3k &  similarity & 0 -- 5\\
STS-16 	& 209 & 868 &  similarity & 0 -- 5\\
\hline
\end{tabular}
\caption{Statistics of the datasets for multi-task learning and the transfer tasks. N is the number of samples, V is the vocabulary size, and C is the number of classes or score range. $^{\dag}$ denotes the datasets that are used in multi-task learning.
}
\label{table:data_stat}
\end{table}


%% file: table/table11.tex
\begin{table*}[ht]
\centering
\small
\begin{tabular}{l c c c c c c}
\hline
\multirow{2}{*}{Model Type} & \multicolumn{2}{c}{QQP} & \multicolumn{2}{c}{SNLI} & \multicolumn{2}{c}{MNLI} \\
& dev & test & dev & test & dev & test \\
\hline 
\multicolumn{7}{l}{Learning from in-domain single task} \\ 
\citep{conneau2017supervised} & 87.1 & 86.7 & 84.7 & 84.5 & 70.2/70.8 & 70.8/69.8  \\
\hline
\multicolumn{7}{l}{Learning from 2-datasets and 2-tasks (SNLI and MNLI)} \\ 
Shared-Private & - & - & 85.0 & {\textbf{85.3}} & {\textbf{71.7/71.4}} & {\textbf{71.8/70.6}} \\
Adversarial Shared-Private & - & - & 84.9 & 84.9 & 70.9/71.4 & 71.0/70.0 \\ 
\hline
\multicolumn{7}{l}{Learning from 2-datasets and 2-tasks (QQP and SNLI)} \\ 
Shared-Private & 87.0 & 86.8 & 84.8 & 84.7 & - & -  \\ 
Adversarial Shared-Private & {87.5} & {\textbf{87.0}} & \textbf{85.2} & {84.9} & - & -    \\ 
\hline
\multicolumn{7}{l}{Learning from 3-datasets and 2-tasks (QQP and AllNLI)} \\ 
Shared-Private & 86.9 & 86.0 & \textbf{85.2} & 84.7 & {70.7/70.5} & {70.8/69.3}  \\ 
Adversarial Shared-Private & 87.0 & 86.6 & 84.7 & 84.3 & 70.2/69.4 & 69.6/68.3 \\ 
\hline
\multicolumn{7}{l}{Learning from 3-datasets and 3-tasks (QQP, SNLI and MNLI)} \\
Shared-Private & {\textbf{87.6}} & {87.0} & \textbf{85.2} & {85.2} & {71.2/71.0} & {71.0/70.1}  \\
Adversarial Shared-Private & 86.6 & 86.3 & 84.6 & 84.7 & 70.7/70.7 & 71.0/70.1  \\ 
\hline
\end{tabular}
\caption{Validation and test accuracy of the source tasks obtained through various multi-task learning architectures. 
Bold-faced values indicate best performance across all the models. 
}
\label{table:source_task_comparison}
\end{table*}

%% file: table/table12.tex
\begin{table*}[ht]
\centering
\resizebox{\linewidth}{!}{%
\begin{tabular}{@{}l@{\hskip 0.05in}| c@{\hskip 0.15in} c@{\hskip 0.15in} c@{\hskip 0.08in} c@{\hskip 0.1in} c@{\hskip 0.1in} c@{\hskip 0.08in} c@{\hskip 0.08in} c@{\hskip 0.05in} c@{\hskip 0.1in} c@{}}
\hline
Model Type & MR & CR & SUBJ & MPQA & SST & TREC & SICK-R & SICK-E & MRPC & STS14\\ 
\hline 
\multicolumn{11}{l}{Sentence representation learning from single-task} \\ 
\hline
BiLSTM-Max (on SNLI)    & 80.1	&85.3	&92.6 & 89.1 & {83.6} & {89.2} & 0.885 & 86.0 & 75.2/82.4  &.66/.64\\ 
BiLSTM-Max (on QQP)	&79.2	& 84.6 & 92.6 & 88.8 & 83.5 & 88.0 & 0.861 & 82.4 & 74.8/82.8	& .62/.60 \\
BiLSTM-Max (on MNLI)   & {81.2}	& 85.8	& 93.1	& {89.5}	& 83.4	& 88.8	& 0.863 & 84.7 	& 75.9/83.1  & .66/.63 \\
BiLSTM-Max (on AllNLI)   & 80.9	& {86.3}	& {93.2}	& 89.2	& 83.3	& 88.8 & {0.887} & {86.7} & {76.4/83.4} & {.69/.66}\\
\hline
\multicolumn{11}{l}{Sentence representation learning from two-tasks (QQP and SNLI)} \\ 
\hline
Shared-Private & 80.5 & 84.8 & {93.4} & 89.1 & {84.0} & 90.2 & 0.881 & 86.1 & 75.1/83.2 & .65/.62 \\
Adversarial Shared-Private & {80.9} & {85.4} & {93.4} & {89.2} & 83.6 & {90.8} & {0.886} & {86.9} & {76.5/82.9} & .68/.65 \\
\hline
\multicolumn{11}{l}{Sentence representation learning from two-tasks (SNLI and MNLI)} \\ 
\hline
Shared-Private & {81.7} &	86.4 & {93.7} & {89.6}	& {84.8}	& 89.2	& 0.885 & 86.7	& 76.3/82.7 & .67/.64 \\
Adversarial Shared-Private & 81.2 &	86.0 &	93.0 & 89.3	& 83.7 & {90.4} & {0.886} & \textbf{{87.1}}	& {76.9/83.5}	& {.70/.67} \\
\hline
\multicolumn{11}{l}{Sentence representation learning from two-tasks (QQP and AllNLI)} \\ 
\hline
Shared-Private & \textbf{{82.0}} & 86.1 & \textbf{{93.9}} & 89.4 & 84.6 & {89.6}	& 0.884 & 86.3 & {76.4/83.4} & .68/.64 \\
Adversarial Shared-Private & 81.4 &	86.3 &	93.2 &	89.4 &	85.1 &	88.4 &	\textbf{{0.888}}	& {86.6}	& 75.5/82.5 & .67/.63 \\
\hline
Shared-Private & 81.6	& {86.9}	& \textbf{{93.9}}	& 89.2	& {84.4}	& 90.4	& 0.883 & 85.9 	&  76.5/83.3		&.66/.63 \\
Adversarial Shared-Private & \textbf{{82.0}} & 86.3 & 93.8 & {89.4} & 84.1 & \textbf{{92.2}} & {0.884} & {87.0} & \textbf{{77.2/83.6}} & .68/.65 \\
\hline 
\end{tabular}
}
\caption{Transfer test results for various single-task and multi-task learning architectures trained on a combination of QQP, SNLI and MNLI datasets. 
Bold-faced values indicate the best performance among all models in this table.}
\label{table:appendix_comparison}
\end{table*}


%% file: table/table7.tex
\begin{table*}[ht]
\centering
\small
\resizebox{\linewidth}{!}{
\begin{tabular}{@{}l@{\hskip 0.1in} c@{\hskip 0.1in} c@{\hskip 0.1in} c@{\hskip 0.1in} c@{\hskip 0.1in} c@{\hskip 0.1in} c@{\hskip 0.1in} c@{\hskip 0.1in} c@{\hskip 0.1in} c@{\hskip 0.1in} c@{}}
\hline
Data Size       & MR  &CR  & SUBJ & MPQA & SST  & TREC & SICK-R & SICK-E & STS14 & MRPC   \\ 
\hline
Same for MTL, STL & $+$0.1 & $+$0.6 & $+$1.6 & $-$0.7 & $+$0.0 & $+$1.7 & $-$0.008 & $-$1.5   & $+$0.7  & $-$0.003 \\
Larger for MTL                  & $+$1.0 & $+$0.8 & $+$1.3 & $-$0.6 & $+$0.3 & $+$2.1 & $+$0.001 & $+$0.6   & $+$1.4  & $+$0.0   \\ \hline
\end{tabular}
}
\caption{The accuracy differences between MTL and STL when training with different sizes of data. For the same data size, MTL and STL are trained on equal amount of annotated data. For larger data size for MTL, MTL is trained on two datasets while STL on one dataset (less data).}
\label{table:wpb}
\end{table*}

%% file: table/table10.tex
\begin{table*}[ht]
\centering
\resizebox{\linewidth}{!}{%
\begin{tabular}{@{}l@{\hskip 0.05in}| c@{\hskip 0.15in} c@{\hskip 0.15in} c@{\hskip 0.08in} c@{\hskip 0.1in} c@{\hskip 0.1in} c@{\hskip 0.08in} c@{\hskip 0.08in} c@{\hskip 0.05in} c@{\hskip 0.1in} c@{}}
\hline
Model Type & MR & CR & SUBJ & MPQA & SST & TREC & SICK-R & SICK-E & MRPC & STS14 \\
\hline
\multicolumn{11}{l}{Shared-Private (trained on SNLI and MNLI)}  \\ 
\hline
Private Encoder (on SNLI) & 79.5 & 84.0 & 92.7 & 89.1 & 82.0 & 87.8 & 0.881 & 84.8 & 75.0/82.7 & .65/.63    \\ 
Private Encoder (on MNLI) & 80.6 & 84.6 & 92.7 & 89.2 & 82.9 & 88.0 & 0.853 & 83.8 & 75.1/82.9 & .60/.58    \\
Shared Encoder & 80.9 & \underline{86.4} & 92.9 & 89.5 & 84.0 & 88.0 & 0.879 & 84.8 & 75.8/83.1 & \underline{.69/.65}    \\
Combined Encoder & \underline{81.7}	& \underline{86.4}	& \underline{93.7}	& \textbf{\underline{89.6}}	& \underline{84.8}	& \underline{89.2}	& \underline{0.885} & \underline{86.7}	& \underline{76.3/82.7}	&.67/.64  \\
\hline
\hline
\multicolumn{11}{l}{Adversarial Shared-Private (trained on SNLI and MNLI)} \\ 
\hline
Private Encoder (on SNLI)     & 79.4 & 84.6 & 92.1 & 89.0 & 82.8  & 86.6  & \underline{0.886}  & 85.5 & 74.0/81.4 & \underline{.68/.66} \\
Private Encoder (on MNLI) & 80.4 & 84.6 & 92.6 & 89.1 & 83.3  & 86.8  & 0.863 & 83.4 & 76.0/83.3 & .65/.63 \\
Shared Encoder                & 81.2     & \underline{86.7}     & 92.4     & 89.3    & 84.5    & 87.0    & 0.875      & 85.1  & 74.8/82.6   & .56/.57        \\
Combined Encoder              & \underline{81.7} & 86.5 & \underline{93.4} & \underline{89.5} & \textbf{\underline{84.9}} & \underline{90.0} & \textbf{\underline{0.888}} & \textbf{\underline{87.1}} & \underline{76.4/83.4} & .64/.63 \\
\hline
\multicolumn{11}{l}{Shared-Private (trained on QQP and AllNLI)}  \\ 
\hline
Private Encoder (on QQP) & 79.4 &	82.6 &	92.5 &	88.5 &	82.2 &	89.2 & 0.856	& 82.9	& 74.3/82.4	& .63/.59    \\
Private Encoder (on AllNLI) & 81.1	& \underline{86.1} &	92.9 &	\underline{89.5} &	83.9	&\underline{90.2}	& 0.876	& 85.0 & 76.0/83.3	& .67/.64    \\
Shared Encoder & 81.2 &	85.6 &	93.1 &	89.2 &	83.4	& 88.2	& 0.880	& 85.3 &	75.7/82.8 & \underline{.69/.66}    \\
Combined Encoder&\textbf{\underline{82.0}}	&\underline{86.1}	&\textbf{\underline{93.9}}	&89.4	&\underline{84.6}	&89.6	&\underline{0.884}	&\underline{86.2}	&\underline{76.4/83.4}	&.68/.64\\
\hline
\multicolumn{11}{l}{Adversarial Shared-Private (trained on QQP and AllNLI)} \\ 
\hline
Private Encoder (on QQP) & 79.2	& 81.7	& 92.1	& 88.7 & 80.8	& 86.6	& 0.865	& 83.8	& 74.1/82.2	& .67/.64 \\
Private Encoder (on AllNLI) & 81.5	& \underline{86.5}	& 92.8 & 89.4 &	82.9 &	88.4	& 0.885	& 85.5 &	75.7/83.2	& \textbf{\underline{.70/.67}}    \\
Shared Encoder & 80.5	& 84.8	& 92.6	& 89.2 &	\underline{83.2}	& 82.6	& 0.876	& 84.8	& 75.5/83.1 & .57/.56   \\
Combined Encoder & \underline{81.9} &	85.9 & 	\underline{93.0}	& \textbf{\underline{89.6}} &	82.4	& \underline{90.6} &	\underline{0.887}	& \underline{86.7}	& \underline{76.8/83.3} & .61/.60   \\
\hline
\multicolumn{11}{l}{Shared-Private (trained on QQP, SNLI and MNLI)}  \\ 
\hline
Private Encoder (on QQP) & 78.9	&83.5	&91.7	&88.3	& 81.4	& 88.6    & 0.850	&82.1	&72.6/81.1	&.62/.59  \\
Private Encoder (on SNLI) &79.1	&83.9	&92.7	&89.0	&81.3	&88.4	   &0.880	&85.2		&74.0/82.0 & .66/.63  \\ 
Private Encoder (on MNLI) &81.0	&85.7	&93.1	&89.3	&82.9	& 89.2	   &0.850	& 84.0		&74.5/82.9  &.60/.58\\ 
Shared Encoder &80.9	&85.8	&92.9	&89.2	&82.7	&85.2	&0.878	&85.8	&76.0/83.1   	& \underline{.68/.65}   \\
Combined Encoder &\underline{81.6}	&\textbf{\underline{86.9}}	&\textbf{\underline{93.9}}	&\underline{89.2}	&\underline{84.4}	& \underline{90.4}	&\underline{0.883}	&\underline{85.9}		&\underline{76.5/83.2}  &.66/.63   \\
\hline
\multicolumn{11}{l}{Adversarial Shared-Private (trained on QQP, SNLI and MNLI)} \\ 
\hline
Private Encoder (on QQP)     & 78.9          & 82.3          & 92.2          & 88.8          & 82.3          & 87.2          & 0.855          & 83.0          & 74.3/82.2          & .64/.62          \\
Private Encoder (on SNLI)      & 79.6          & 84.4          & 92.0          & 89.0          & 82.7          & 88.2          & 0.881          & 85.4          & 74.6/82.4          & \underline{.67/.65} \\
Private Encoder (on MNLI) & 80.6          & 84.7          & 93.0          & 89.1          & 83.6          & 89.2          & 0.863          & 84.8          & 75.8/82.8          & .65/.62          \\
Shared Encoder                 & 81.0          & 85.9          & 92.7          & \textbf{\underline{89.6}} & 82.9          & 87.0          & 0.876          & 85.7          & 74.8/82.8          & .66/.64          \\
Combined Encoder               &\textbf{\underline{82.0}} & \underline{86.3} & \underline{93.8} & 89.4          & \underline{84.1} & \textbf{\underline{92.2}} & \underline{0.884} & \underline{87.0} & \textbf{\underline{77.2/83.6}} & .66/.64 \\
\hline
\end{tabular}
}
\caption{Detailed analysis of the transfer test results for shared-private models trained on different combinations of QQP, SNLI and MNLI datasets. Combined encoder refers to the concatenation of shared encoder and all private encoders. Underlined values indicate the best performance among different encoders of the shared-private models trained on the same set of tasks. Bold-faced values indicate the best performance among all models in this table.}
\label{table:sp_comparison}
\end{table*}